\documentclass[conference, letterpaper]{ieeeconf}
\usepackage[latin9]{inputenc}
\usepackage{color}
\usepackage{url}
\usepackage{amsmath}
\usepackage{amssymb}
\usepackage[unicode=true,
 bookmarks=true,bookmarksnumbered=true,bookmarksopen=true,bookmarksopenlevel=1,
 breaklinks=false,pdfborder={0 0 1},backref=false,colorlinks=false]
 {hyperref}
\hypersetup{pdftitle={Your Title},
 pdfauthor={Your Name},
 dvipdfmx,setpagesize=false,pdfborderstyle=,pdfpagelayout=OneColumn,pdfnewwindow=true,pdfstartview=XYZ,plainpages=false}

\makeatletter
\IEEEoverridecommandlockouts
\usepackage{cite}
\usepackage{textcomp}
\usepackage{xcolor}
\usepackage[caption=false,font=footnotesize]{subfig}
\usepackage{resizegather}
\usepackage{import}
\usepackage{graphicx}

\makeatother

\begin{document}
\global\long\def\quat#1{\boldsymbol{#1}}%

\global\long\def\dq#1{\underline{\boldsymbol{#1}}}%

\global\long\def\hp{\mathbb{H}_{p}}%

\global\long\def\dotmul#1#2{\left\langle #1,#2\right\rangle }%

\global\long\def\partialfrac#1#2{\frac{\partial\left(#1\right)}{\partial#2}}%

\global\long\def\totalderivative#1#2{\frac{d}{d#2}\left(#1\right)}%

\global\long\def\mymatrix#1{\boldsymbol{#1}}%

\global\long\def\vecthree#1{\operatorname{v}_{3}\left(#1\right)}%

\global\long\def\vecfour#1{\operatorname{v}_{4}\left(#1\right)}%

\global\long\def\haminuseight#1{\overset{-}{\mymatrix H}_{8}\left(#1\right)}%

\global\long\def\hapluseight#1{\overset{+}{\mymatrix H}_{8}\left(#1\right)}%

\global\long\def\haminus#1{\overset{-}{\mymatrix H}_{4}\left(#1\right)}%

\global\long\def\haplus#1{\overset{+}{\mymatrix H}_{4}\left(#1\right)}%

\global\long\def\norm#1{\left\Vert #1\right\Vert }%

\global\long\def\abs#1{\left|#1\right|}%

\global\long\def\conj#1{#1^{*}}%

\global\long\def\veceight#1{\operatorname{v}_{8}\left(#1\right)}%

\global\long\def\myvec#1{\boldsymbol{#1}}%

\global\long\def\imi{\hat{\imath}}%

\global\long\def\imj{\hat{\jmath}}%

\global\long\def\imk{\hat{k}}%

\global\long\def\dual{\varepsilon}%

\global\long\def\getp#1{\operatorname{\mathcal{P}}\left(#1\right)}%

\global\long\def\getpdot#1{\operatorname{\dot{\mathcal{P}}}\left(#1\right)}%

\global\long\def\getd#1{\operatorname{\mathcal{D}}\left(#1\right)}%

\global\long\def\getddot#1{\operatorname{\dot{\mathcal{D}}}\left(#1\right)}%

\global\long\def\real#1{\operatorname{\mathrm{Re}}\left(#1\right)}%

\global\long\def\imag#1{\operatorname{\mathrm{Im}}\left(#1\right)}%

\global\long\def\spin{\text{Spin}(3)}%

\global\long\def\spinr{\text{Spin}(3){\ltimes}\mathbb{R}^{3}}%

\global\long\def\distance#1#2#3{d_{#1,\mathrm{#2}}^{#3}}%

\global\long\def\distancejacobian#1#2#3{\boldsymbol{J}_{#1,#2}^{#3}}%

\global\long\def\distancegain#1#2#3{\eta_{#1,#2}^{#3}}%

\global\long\def\distanceerror#1#2#3{\tilde{d}_{#1,#2}^{#3}}%

\global\long\def\dotdistance#1#2#3{\dot{d}_{#1,#2}^{#3}}%

\global\long\def\distanceorigin#1{d_{#1}}%

\global\long\def\dotdistanceorigin#1{\dot{d}_{#1}}%

\global\long\def\squaredistance#1#2#3{D_{#1,#2}^{#3}}%

\global\long\def\dotsquaredistance#1#2#3{\dot{D}_{#1,#2}^{#3}}%

\global\long\def\squaredistanceerror#1#2#3{\tilde{D}_{#1,#2}^{#3}}%

\global\long\def\squaredistanceorigin#1{D_{#1}}%

\global\long\def\dotsquaredistanceorigin#1{\dot{D}_{#1}}%

\global\long\def\crossmatrix#1{\overline{\mymatrix S}\left(#1\right)}%

\global\long\def\constraint#1#2#3{\mathcal{C}_{\mathrm{#1},\mathrm{#2}}^{\mathrm{#3}}}%

\global\long\def\si{\text{R1}}%

\global\long\def\lg{\text{R2}}%

\global\long\def\la{a}%

\global\long\def\vecc{\quat p_{c}^{\lg}}%

\global\long\def\dotvecc{\dot{\quat p}_{c}^{\lg}}%

\global\long\def\javecc{\boldsymbol{J}_{c}^{\lg}}%

\global\long\def\vece{\quat p_{e}^{\lg}}%

\global\long\def\dotvece{\dot{\quat p}_{e}^{\lg}}%

\global\long\def\veca{\quat p_{\la}^{\lg}}%

\global\long\def\dotveca{\dot{\quat p}_{\la}^{\lg}}%

\global\long\def\javeca{\boldsymbol{J}_{\la}^{\lg}}%

\global\long\def\vecasi{\quat p_{\si}^{\la}}%

\global\long\def\dotvecasi{\dot{\quat p}_{\si}^{\la}}%

\global\long\def\javecasi{\boldsymbol{J}_{\si}^{\la}}%

\global\long\def\vecsi{\quat t_{\si}^{\lg}}%

\global\long\def\dotvecsi{\dot{\quat t}_{\si}^{\lg}}%

\global\long\def\javecsi{\boldsymbol{J}_{\si}^{\lg}}%

\global\long\def\fworld{\mathcal{F}_{\text{W}}}%

\global\long\def\flg{\mathcal{F}_{\lg}}%

\global\long\def\centerp{\quat p_{c}}%

\global\long\def\edgep{\quat p_{e}}%

\global\long\def\sip{\quat t_{1}}%

\global\long\def\lgp{\quat t_{2}}%

\global\long\def\sipd{\quat t_{1,\text{d}}}%

\global\long\def\lgpd{\quat t_{2,\text{d}}}%

\global\long\def\dotsip{\dot{\quat t}_{1}}%

\global\long\def\dotlgp{\dot{\quat t}_{2}}%

\global\long\def\dotqsi{\dot{\quat q}_{1}}%

\global\long\def\dotqlg{\dot{\quat q}_{2}}%

\global\long\def\dotq{\dot{\quat q}}%

\global\long\def\qsi{\quat q_{1}}%

\global\long\def\qlg{\quat q_{2}}%

\global\long\def\transjasi{\boldsymbol{J}_{t_{1}}}%

\global\long\def\transjalg{\boldsymbol{J}_{t_{2}}}%

\global\long\def\rotatejalg{\boldsymbol{J}_{r_{2}}}%

\global\long\def\cscope{\text{C1}}%

\global\long\def\cillu{\text{C2}}%

\global\long\def\disjascope{\mymatrix J_{d,\cscope}}%

\global\long\def\disjaillu{\mymatrix J_{d,\cillu}}%

\global\long\def\disjascopesafe{\mymatrix J_{d,\cscope,\text{safe}}}%

\global\long\def\disjaillusafe{\mymatrix J_{d,\cillu,\text{safe}}}%

\global\long\def\disscope{d_{\cscope}}%

\global\long\def\disillu{d_{\cillu}}%

\global\long\def\dotdisscope{\dot{d}_{\cscope}}%

\global\long\def\dotdisillu{\dot{d}_{\cillu}}%

\global\long\def\disscopesafe{d_{\cscope,\text{safe}}}%

\global\long\def\disillusafe{d_{\cillu,\text{safe}}}%

\global\long\def\dotdisscopesafe{\dot{d}_{\cscope,\text{safe}}}%

\global\long\def\dotdisillusafe{\dot{d}_{\cillu,\text{safe}}}%

\global\long\def\wsradius{r}%

\global\long\def\thetascope{\theta_{\cscope}}%

\global\long\def\thetascopesafe{\theta_{\cscope,\text{safe}}}%

\global\long\def\thetaillu{\theta_{\cillu}}%

\global\long\def\thetaillusafe{\theta_{\cillu,\text{safe}}}%

\global\long\def\o{\boldsymbol{0}}%

\global\long\def\lgdir{\quat l_{2}}%

\global\long\def\lgr{\quat r_{2}}%

\global\long\def\dotlgdir{\dot{\quat l}_{2}}%

\global\long\def\dotlgr{\dot{\quat r}_{2}}%

\global\long\def\shaftdis{d_{\text{shaft}}}%

\global\long\def\tipdis{d_{\text{tip}}}%

\global\long\def\safeineqleft{\boldsymbol{W}_{\text{safe}}}%

\global\long\def\safeineqright{\boldsymbol{w}_{\text{safe}}}%

\global\long\def\illuineqleft{\boldsymbol{W}_{\text{lg}}}%

\global\long\def\illuineqright{\boldsymbol{w}_{\text{lg}}}%

\global\long\def\eyeballineqleft{\boldsymbol{W}_{\text{eyeball}}}%

\global\long\def\eyeballineqright{\boldsymbol{w}_{\text{eyeball}}}%

\global\long\def\cscope{\text{C}_{\text{sh}}}%

\global\long\def\scope{\text{sh}}%

\global\long\def\cillu{\text{C}_{\text{i}}}%

\global\long\def\illu{\text{i}}%

\global\long\def\csecond{\text{C3}}%

\global\long\def\rcm{\text{\text{R}}}%

\global\long\def\crcm{\text{C}_{\text{\ensuremath{\rcm}}}}%

\global\long\def\newcrcm{\text{C}_{\text{\ensuremath{\rcm}}}^{'}}%

\global\long\def\retina{\text{\text{r}}}%

\global\long\def\ceyeball{\text{C}_{\retina}}%

\global\long\def\shaft{\text{s}}%

\global\long\def\cshaft{\text{C}_{\shaft}}%

\global\long\def\trocar{\text{tr}}%

\global\long\def\ctrocar{\text{C}_{\trocar}}%

\global\long\def\newctrocar{\text{C}_{\trocar}^{'}}%

\global\long\def\tip{\text{\text{t}}}%

\global\long\def\ctip{\text{C}_{\tip}}%

\global\long\def\micro{\text{\text{m}}}%

\global\long\def\cmicro{\text{C}_{\micro}}%

\global\long\def\robot{\text{ro}}%

\global\long\def\crobot{\text{C}_{\robot}}%

\global\long\def\joint{\text{j}}%

\global\long\def\cjoint{\text{C}_{\joint}}%

\global\long\def\worldcoordinate{o}%

\global\long\def\worldreferenceframe{\mathcal{F}_{\text{\ensuremath{\worldcoordinate}}}}%

\global\long\def\eyeballcoordinate{e}%

\global\long\def\eyeballreferenceframe{\mathcal{F}_{\text{\ensuremath{\eyeballcoordinate}}}}%

\global\long\def\eyeballmediatecoordinate{m}%

\global\long\def\eyeballmediatereferenceframe{\mathcal{F}_{\text{\ensuremath{\eyeballmediatecoordinate}}}}%

\global\long\def\orbital{\text{OM}}%

\global\long\def\si{\text{R1}}%

\global\long\def\lg{\text{R2}}%

\global\long\def\generalrobot{\text{R}i}%

\global\long\def\eyeballerror{\tilde{\quat t}_{e}}%

\global\long\def\doteyeballerror{\tilde{\dot{\quat t}}_{e}}%

\global\long\def\jaeyeball{\boldsymbol{J}_{\text{eyeball}}}%

\global\long\def\eyeballtd{\quat t_{\si,\text{d}}}%

\global\long\def\doteyeballtd{\dot{\quat t}_{\si,\text{d}}}%

\global\long\def\eyeballreferencetd{\quat t_{\si,\text{d}}^{e}}%

\global\long\def\jad{\boldsymbol{J}_{\eyeballtd}}%

\global\long\def\sit{\quat t_{\text{1}}}%

\global\long\def\lgt{\quat t_{2}}%

\global\long\def\generalt{\quat t_{i}}%

\global\long\def\dotsit{\dot{\quat t}_{1}}%

\global\long\def\dotlgt{\dot{\quat t}_{2}}%

\global\long\def\dotgeneralt{\dot{\quat t}_{i}}%

\global\long\def\transjasi{\boldsymbol{J}_{\sit}}%

\global\long\def\transjalg{\boldsymbol{J}_{\lgt}}%

\global\long\def\transjageneral{\boldsymbol{J}_{\generalt}}%

\global\long\def\generalrcmt{\quat t_{\orbital,i}}%

\global\long\def\sircmt{\quat t_{\orbital,1}}%

\global\long\def\lgrcmt{\quat t_{\orbital,2}}%

\global\long\def\dotsircmt{\dot{\quat t}_{\orbital,1}}%

\global\long\def\dotlgrcmt{\dot{\quat t}_{\orbital,2}}%

\global\long\def\dotgeneralrcmt{\dot{\quat t}_{\orbital,i}}%

\global\long\def\sircmtmediate{\quat t_{\text{R}_{\si},\text{init}}^{\eyeballmediatecoordinate}}%

\global\long\def\lgrcmtmediate{\quat t_{\text{R}_{\lg}}^{\eyeballmediatecoordinate}}%

\global\long\def\lgrcmtmed{\quat t_{\text{R}_{\lg},\text{init}}^{\eyeballmediatecoordinate}}%

\global\long\def\sircmtinit{\quat t_{\text{R}_{\si},\text{init}}^{\worldcoordinate}}%

\global\long\def\lgrcmtinit{\quat t_{\text{R}_{\lg},\text{init}}^{\worldcoordinate}}%

\global\long\def\jarcmtgeneral{\boldsymbol{J}_{\generalrcmt}}%

\global\long\def\jasircmt{\boldsymbol{J}_{\sircmt}}%

\global\long\def\jalgrcmt{\boldsymbol{J}_{\lgrcmt}}%

\global\long\def\eyet{\quat t_{e}}%

\global\long\def\sir{\quat r_{1}}%

\global\long\def\lgr{\quat r_{2}}%

\global\long\def\generalr{\quat r_{i}}%

\global\long\def\dotsir{\dot{\quat r}_{1}}%

\global\long\def\dotlgr{\dot{\quat r}_{2}}%

\global\long\def\generaldotr{\dot{\quat r}_{i}}%

\global\long\def\jasir{\boldsymbol{J}_{\sir}}%

\global\long\def\jalgr{\boldsymbol{J}_{\lgr}}%

\global\long\def\rotationjageneral{\boldsymbol{J}_{\generalr}}%

\global\long\def\silength{d_{1}}%

\global\long\def\lglength{d_{2}}%

\global\long\def\generallength{d_{i}}%

\global\long\def\dotsilength{\dot{d}_{1}}%

\global\long\def\dotlglength{\dot{d}_{2}}%

\global\long\def\dotgenerallength{\dot{d}_{i}}%

\global\long\def\jasilength{\boldsymbol{J}_{\silength}}%

\global\long\def\jalglength{\boldsymbol{J}_{\lglength}}%

\global\long\def\jagenerallength{\boldsymbol{J}_{\generallength}}%

\global\long\def\eyer{\quat r_{e}}%

\global\long\def\doteyer{\dot{\quat r}_{e}}%

\global\long\def\jaeyer{\boldsymbol{J}_{\eyer}}%

\global\long\def\thetafirst{\quat{\theta}_{1}}%

\global\long\def\thetasecond{\quat{\theta}_{2}}%

\global\long\def\eyerfirst{\quat r_{\eyeballmediatecoordinate}^{\worldcoordinate}}%

\global\long\def\doteyerfirst{\dot{\quat r}_{\eyeballmediatecoordinate}^{\worldcoordinate}}%

\global\long\def\jaeyerfirst{\boldsymbol{J}_{r_{\eyeballmediatecoordinate}^{\worldcoordinate}}}%

\global\long\def\eyersecond{\quat r_{e}^{\eyeballmediatecoordinate}}%

\global\long\def\doteyersecond{\dot{\quat r}_{e}^{\eyeballmediatecoordinate}}%

\global\long\def\jaeyersecond{\boldsymbol{J}_{r_{e}^{\eyeballmediatecoordinate}}}%

\global\long\def\eyerotationaxisfirst{\quat n^{\worldcoordinate}}%

\global\long\def\eyerotationaxissecond{\quat n^{\eyeballmediatecoordinate}}%

\global\long\def\sircminitunit{\quat u_{\text{R}_{\si},\text{init}}^{\worldcoordinate}}%

\global\long\def\lgrcminitunit{\quat u_{\text{R}_{\lg},\text{init}}^{\worldcoordinate}}%

\global\long\def\sircmunit{\quat u_{\text{R}_{\si}}^{\worldcoordinate}}%

\global\long\def\crossfirst{\quat a_{1}}%

\global\long\def\innerprofirst{a_{2}}%

\global\long\def\athree{\quat A_{3}}%

\global\long\def\afour{\boldsymbol{A}_{4}}%

\global\long\def\afive{\boldsymbol{A}_{5}}%

\global\long\def\crosssecond{\quat b_{1}}%

\global\long\def\innerprosecond{b_{2}}%

\global\long\def\bthree{\quat B_{3}}%

\global\long\def\bfour{\quat B_{4}}%

\global\long\def\bfive{\quat B_{5}}%

\global\long\def\cfirst{\quat c_{1}}%

\global\long\def\csecond{\quat C_{2}}%

\global\long\def\cthird{\quat C_{3}}%

\global\long\def\dotsircmunit{\dot{\quat u}_{\text{R}_{\si}}^{\worldcoordinate}}%

\global\long\def\jasircmunit{\boldsymbol{J}_{\sircmunit}}%

\global\long\def\lgrcmunit{\quat u_{\text{R}_{\lg}}^{\worldcoordinate}}%

\global\long\def\dotlgrcmunit{\dot{\quat u}_{\text{R}_{\lg}}^{\worldcoordinate}}%

\global\long\def\jalgrcmunit{\boldsymbol{J}_{\lgrcmunit}}%

\global\long\def\sircmunitsecond{\quat u_{\text{R}_{\si},\text{init}}^{\eyeballmediatecoordinate}}%

\global\long\def\lgrcmunitsecond{\quat u_{\text{R}_{\lg}}^{\eyeballmediatecoordinate}}%

\global\long\def\lgrcmunitseconddash{\quat u_{\text{R}_{\lg},\text{init}}^{\eyeballmediatecoordinate}}%

\global\long\def\dotlgrcmunitsecond{\dot{\quat u}_{\text{R}_{\lg}}^{\eyeballmediatecoordinate}}%

\global\long\def\jalgrcmunitsecond{\boldsymbol{J}_{\lgrcmunitsecond}}%

\global\long\def\lgrcmunitperpendicular{\quat u_{\text{h}}^{\eyeballmediatecoordinate}}%

\global\long\def\dotlgrcmunitperpendicular{\dot{\quat u}_{\text{h}}^{\eyeballmediatecoordinate}}%

\global\long\def\jalgrcmunitperpendicular{\boldsymbol{J}_{\lgrcmunitperpendicular}}%

\global\long\def\lgrcmunitperpendiculardash{\quat u_{\text{h,init}}^{\eyeballmediatecoordinate}}%

\global\long\def\dotlgrcmunitperpendiculardash{\dot{\quat u}_{3}^{\eyeballmediatecoordinate'}}%

\global\long\def\jalgrcmunitperpendiculardash{\boldsymbol{J}_{\lgrcmunitperpendiculardash}}%

\global\long\def\orthgraphic{\quat h^{\eyeballmediatecoordinate}}%

\global\long\def\eyeballradius{r_{\text{eye}}}%

\global\long\def\dotqsi{\dot{\quat q}_{1}}%

\global\long\def\dotqlg{\dot{\quat q}_{2}}%

\global\long\def\dotq{\dot{\quat q}}%

\global\long\def\generalq{\quat q_{i}}%

\global\long\def\generalqdot{\dot{\quat q}_{i}}%

\global\long\def\generalline{\quat l_{i}}%

\global\long\def\siline{\quat l_{1}}%

\global\long\def\lgline{\quat l_{2}}%

\global\long\def\dotgeneralline{\dot{\quat l}_{i}}%

\global\long\def\dotsiline{\dot{\quat l}_{1}}%

\global\long\def\dotlgline{\dot{\quat l}_{2}}%

\global\long\def\jageneralline{\boldsymbol{J}_{\generalline}}%

\global\long\def\zeromatrix{\boldsymbol{O}}%

\global\long\def\sijointnumber{n_{1}}%

\global\long\def\lgjointnumber{n_{2}}%

\global\long\def\cfour{\boldsymbol{C}_{4}}%

\global\long\def\fworld{\mathcal{F}_{\text{W}}}%

\global\long\def\flg{\mathcal{F}_{\lg}}%

\global\long\def\rcmdplus{\tilde{D}_{\orbital}^{+}}%

\global\long\def\rcmdminus{\tilde{D}_{\orbital}^{-}}%

\global\long\def\dotrcmdplus{\dot{\tilde{D}}_{\orbital}^{+}}%

\global\long\def\dotrcmdminus{\dot{\tilde{D}}_{\orbital}^{-}}%

\global\long\def\jarcmdplus{\boldsymbol{J}_{\orbital}}%

\global\long\def\jarcmdminus{\boldsymbol{J}_{\orbital}}%

\global\long\def\rotationplanefirst{\quat{\pi}_{\rotation1}}%

\global\long\def\rotationplanesecond{\quat{\pi}_{\rotation2}}%

\global\long\def\rotation{\text{rot}}%

\global\long\def\crotation{\text{C}_{\rotation j}}%

\global\long\def\crotationfirst{\text{C}_{\rotation1}}%

\global\long\def\crotationsecond{\text{C}_{\rotation2}}%

\global\long\def\rotgeneraldfirst{d_{\rotation j,1}}%

\global\long\def\rotfirstdfirst{d_{\rotation1,1}}%

\global\long\def\rotfirstdsecond{d_{\rotation1,2}}%

\global\long\def\rotgeneraldsecond{d_{\rotation j,2}}%

\global\long\def\rotseconddfirst{d_{\rotation2,1}}%

\global\long\def\rotseconddsecond{d_{\rotation2,2}}%

\global\long\def\rotplaned{d_{\rotationplanesecond}}%

\global\long\def\rotgeneraljafirst{\boldsymbol{J}_{\rotation j,1}}%

\global\long\def\rotfirstjafirst{\boldsymbol{J}_{\rotation1,1}}%

\global\long\def\rotfirstjasecond{\boldsymbol{J}_{\rotation1,2}}%

\global\long\def\rotgeneraljasecond{\boldsymbol{J}_{\rotation j,2}}%

\global\long\def\rotsecondjafirst{\boldsymbol{J}_{\rotation2,1}}%

\global\long\def\rotsecondjasecond{\boldsymbol{J}_{\rotation2,2}}%

\global\long\def\trocarjanewfirst{\boldsymbol{J}_{\trocar,1}^{'}}%

\global\long\def\trocarjanewsecond{\boldsymbol{J}_{\trocar,2}^{'}}%

\global\long\def\trocarjanewgeneral{\boldsymbol{J}_{\trocar,i}^{'}}%

\global\long\def\manipulability{\text{}}%

\global\long\def\manija{\boldsymbol{J}}%

\global\long\def\manijaoldrcm{\boldsymbol{J}_{\text{w/o}}}%

\global\long\def\manijanewrcm{\boldsymbol{J}_{\text{with}}}%

\global\long\def\simulationrcmjasi{\boldsymbol{J}_{\crcm,\si}}%

\global\long\def\simulationrcmjalg{\boldsymbol{J}_{\crcm,\lg}}%

\global\long\def\simulationrcmjageneral{\boldsymbol{J}_{\crcm,\generalrobot}}%

\global\long\def\simulationrcmjanew{\boldsymbol{J}_{\orbital}^{'}}%

\global\long\def\manimeasure{\boldsymbol{\omega}}%

\global\long\def\manimeasureold{\boldsymbol{\omega}_{\crcm}}%

\global\long\def\manimeasurenew{\boldsymbol{\omega}_{\newcrcm}}%

\title{Vitreoretinal Surgical Robotic System with Autonomous Orbital Manipulation
using Vector-Field Inequalities}
\author{Yuki~Koyama, Murilo~M.~Marinho, and Kanako~Harada\thanks{This
research was funded in part by the ImPACT Program of the Council for
Science, Technology and Innovation (Cabinet Office, Government of
Japan), and in part by the Mori Manufacturing Research and Technology
Foundation.}\thanks{(\emph{Corresponding author:} Murilo~M.~Marinho)}\thanks{Yuki~Koyama,
Murilo~M.~Marinho, and Kanako~Harada are with the Department of
Mechanical Engineering, the University of Tokyo, Tokyo, Japan. \texttt{Emails:\{yuuki-koyama581,
murilo, kanakoharada\}}@g.ecc.u-tokyo.ac.jp. }}
\maketitle
\begin{abstract}
Vitreoretinal surgery pertains to the treatment of delicate tissues
on the fundus of the eye using thin instruments. Surgeons frequently
rotate the eye during surgery, which is called orbital manipulation,
to observe regions around the fundus without moving the patient. In
this paper, we propose the autonomous orbital manipulation of the
eye in robot-assisted vitreoretinal surgery with our tele-operated
surgical system. In a simulation study, we preliminarily investigated
the increase in the manipulability of our system using orbital manipulation.
Furthermore, we demonstrated the feasibility of our method in experiments
with a physical robot and a realistic eye model, showing an increase
in the view-able area of the fundus when compared to a conventional
technique. \textcolor{blue}{Source code and minimal example available
at \url{https://github.com/mmmarinho/icra2023_orbitalmanipulation}.}
\end{abstract}

\section{Introduction}

Vitreoretinal surgery is among the most challenging microsurgeries.
Tasks require a high level of manipulation skill with two surgical
instruments; a dominant instrument, such as a 0.5 mm (25 gauge) diameter
forceps or needle, and a light guide to illuminate the workspace.
With these instruments, surgeons need to peel off $2.5$ $\mathrm{\mu m}$
thick inner-limiting membranes or insert a needle into $100$ $\mathrm{\mu m}$
diameter retinal blood vessels. Moreover, hand tremors with an average
amplitude of approximately $100\mathrm{\,\mu m}$ \cite{singhPhysiologicalTremorAmplitude2002}
make these procedures more difficult.

To address these difficulties, several robotic systems have been developed
\cite{iordachitaRoboticAssistanceIntraocular2022}. These systems
can be classified into three fundamentally different approaches: hand-held
robotic devices \cite{maclachlanMicronActivelyStabilized2012,kimDesignControlFully2021},
cooperatively-controlled systems \cite{uneriNewSteadyhandEye2010,gijbelsClinicallyApplicableRobotic2016},
and tele-operated systems \cite{weiDesignTheoreticalEvaluation2007,nasseriIntroductionNewRobot2013,wilsonIntraocularRoboticInterventional2018}.
Furthermore, some systems were already used in clinical settings \cite{edwardsFirstinhumanStudySafety2018a,gijbelsInHumanRobotAssistedRetinal2018}.

\begin{figure}[t]
\centering
\def\svgwidth{240pt} 
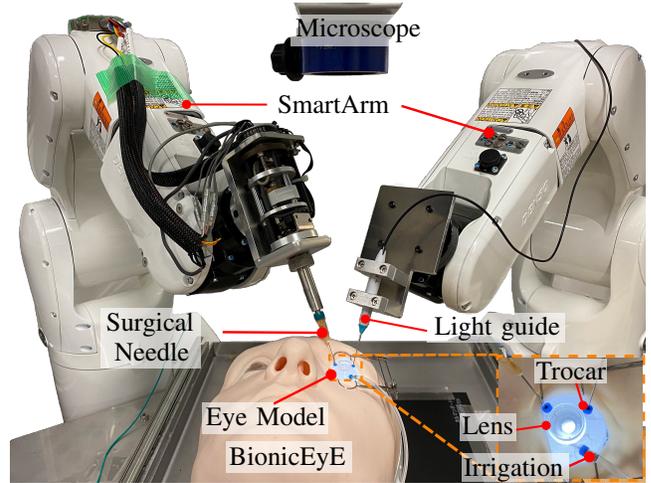

\caption{\label{fig:surgical_system} The vitreoretinal setup of the SmartArm
robotic system \cite{marinhoSmartArmIntegrationValidation2020}. The
experiments described in Section \ref{subsec:Experiment} were conducted
without the face, as it was not made to consider orbital manipulation.}
\end{figure}

Most robotic systems generate, either through hardware or software
constraints, a fixed \emph{remote center-of-motion} (RCM). RCMs are
required for the instruments to move through incisions and prevent
damage to the patient. In surgical practice, e.g., the peeling of
a specific area of the membrane on the fundus, the ophthalmic surgeons
routinely and purposefully move the RCMs of each instrument, which
is called \emph{orbital manipulation}. The human eye has the freedom
to rotate around its center, and surgeons take advantage of orbital
manipulation to view and approach specific areas in the eye without
moving the patient. Despite this widespread use in surgical practice,
notably, only the group of \cite{weiDesignTheoreticalEvaluation2007,weiweiPerformanceEvaluationMultiarm2009},
and \cite{yuDesignCalibrationPreliminary2013} has addressed this
topic\footnote{To the best of the authors' knowledge, the term \emph{orbital manipulation}
was first used in a robotics context in \cite{weiDesignTheoreticalEvaluation2007}.}.

In our group, we have been developing a versatile surgical robotic
system for constrained workspaces, called the SmartArm surgical robotic
system. The SmartArm system has already been validated using realistic
phantoms in some types of surgery \cite{marinhoSmartArmIntegrationValidation2020,marinhoUnifiedFrameworkTeleoperation2019}
and has shown to have enough accuracy for vitreoretinal procedures
\cite{tomikiUseGeneralpurposeSeriallink2017}. The setup of the SmartArm
system for vitreoretinal surgery is shown in Fig.~\ref{fig:surgical_system}.

To improve the safety and efficiency of robot-assisted vitreoretinal
surgery, we have been working on automation \cite{koyamaAutonomousCoordinatedControl2022}.
Our technique also naturally allows for the semi-autonomous scenario
where the surgeon tele-operates one of the instruments, and the system
autonomously controls the other one, which is a light guide. This
semi-autonomous scenario has the potential to improve the efficiency
of surgical procedures and possibly lead to new surgical techniques
by setting the one hand of the surgeon free.

In this work, we take our semi-autonomous system one step further
and add autonomous orbital manipulation. By letting the eye autonomously
rotate with respect to the motion of the instruments, surgeons can
perform vitreoretinal tasks in a wider workspace without moving the
patient.

\subsection{Related works}

The automation of vitreoretinal tasks is an active research field.
In recent years, He \emph{et al.} \cite{heAutomaticLightPipe2020}
performed the bimanual control of instruments and proposed an automatic
light pipe actuation system. Kim \emph{et al. }\cite{kimAutonomousEyeSurgery2021}
automated a tool-navigation task using deep imitation learning. Shin
\emph{et al. }\cite{shinSemiAutomatedExtractionLens2021} tackled
the semi-autonomous extraction of lens fragments. Dehghani \emph{et
al.} \cite{dehghaniColibriDocEyeInHandAutonomous2022} achieved autonomous
docking of the instrument to the trocar.

Regarding object manipulation, some works addressed rigid multibody
systems \cite{leeDualRedundantArm1989,wenKinematicManipulabilityGeneral1999a}
and collaborative deformable tissue manipulation \cite{alambeigiSemiautonomousCryoablationKidney2018}.
However, these strategies cannot be directly applied to orbital manipulation,
where the instruments penetrate the eyeball instead of grabbing it.

To the best of the authors' knowledge, the works closest to our objectives
are the ones from Wei \emph{et al.} \cite{weiweiPerformanceEvaluationMultiarm2009},
where they proposed the mathematical model of orbital manipulation
by separating intraocular manipulation from orbital manipulation,
and Yu \emph{et al.} \cite{yuDesignCalibrationPreliminary2013} from
the same group, where they demonstrated their method with a physical
model.

In their mathematical model, effective for their purposes, intraocular
and orbital manipulation were controlled individually. Our purpose
is, instead, to automate orbital manipulation in a transparent way
without any change in the description of the task. This autonomous
orbital manipulation also requires the definition of hard limits for
the motion of the eye, which cannot be achieved by prior work. In
this context, we propose a new control strategy for orbital manipulation
using inequality constraints based on the vector-field inequalities
(VFIs) methodology \cite{marinhoDynamicActiveConstraints2019}.

\subsection{Statement of contributions\label{subsec:Statement-of-contributions}}

The main contributions of this work are:
\begin{enumerate}
\item new VFIs for orbital manipulation based on new distance functions
and Jacobians unique to this task, and
\item simulation and experimental results in a physical robotic system to
evaluate the effects of orbital manipulation and its feasibility.
\end{enumerate}

\section{Problem Statement}

Fig.~\ref{fig:surgical_system} shows the vitreoretinal surgical
robotic setup of the SmartArm system. In this work, a surgical needle
$\left(25\,\mathrm{G}\right)$ was used as a dominant surgical instrument.
Let the robot that holds the surgical needle be $\si$, with joint
values $\qsi\in\mathbb{R}^{1\times\sijointnumber}$, and the robot
that holds the light guide be $\lg$, with joint values $\qlg\in\mathbb{R}^{1\times\lgjointnumber}$.
A vitreoretinal surgical phantom with realistic physical properties
(BionicEyE \cite{omataSurgicalSimulatorPeeling2018a}) is placed between
the two robots. The instruments are inserted into the eye model through
ophthalmic trocars. As in surgical practice, a disposable flat lens
is placed on the eye model to provide the correct view of the workspace.
Images of the workspace are obtained through an ophthalmic microscope
placed above the BionicEyE.

We have already considered various requirements unique to robot-assisted
vitreoretinal surgery \cite{koyamaAutonomousCoordinatedControl2022}.
In this work, our goal is to autonomously perform orbital manipulation
to view unseen areas on the fundus without moving the patient.

\section{Mathematical background}

In this section, we summarize the required background in quaternion
algebra, constrained optimization, and VFIs.

\subsection{Quaternions and operators\label{subsec:Quaternions-and-operators}}

The quaternion set is
\begin{align*}
\mathbb{H}\triangleq & \left\{ h_{1}+\imi h_{2}+\imj h_{3}+\imk h_{4}\,:\,h_{1},h_{2},h_{3},h_{4}\in\mathbb{R}\right\} ,
\end{align*}
where $\hat{\imath}^{2}=\hat{\jmath}^{2}=\hat{k}^{2}=\hat{\imath}\hat{\jmath}\hat{k}=-1$.
Elements of the set $\mathbb{H}_{p}\triangleq\left\{ \quat h\in\mathbb{H}\,:\,\real{\quat h}=0\right\} $
represent translations in $\mathbb{R}^{3}$. The set of quaternions
with unit norm, $\mathbb{S}^{3}\triangleq\left\{ \quat r\in\mathbb{H}\,:\,\norm{\quat r}=1\right\} $,
represent rotations. We use the operator $\text{\ensuremath{\mathrm{v}_{4}}}$
to map a quaternion $\quat h$ $\in$ $\mathbb{H}$ into a column
vector $\mathbb{R}^{4}$. Moreover, the Hamilton operators $\overset{+}{\mymatrix H}_{4}$
and $\overset{-}{\mymatrix H}_{4}$ \cite[Def. 2.1.6]{adornoRobotKinematicModeling2017}
satisfy $\vecfour{\quat h\quat h^{\prime}}=\haplus{\quat h}\vecfour{\quat h^{\prime}}=\haminus{\quat h^{\prime}}\vecfour{\quat h}$,
$\mymatrix C_{4}=\text{diag}(1,\,-1,\,-1,\,-1)$ satisfies $\vecfour{\conj{\quat h}}=\mymatrix C_{4}\vecfour{\quat h}$,
and $\overline{\mymatrix S}$ \cite[Eq. (3)]{marinhoDynamicActiveConstraints2019}
has the properties $\vecfour{\quat h\times\quat h^{\prime}}=\crossmatrix{\quat h}\vecfour{\quat h^{\prime}}=\crossmatrix{\quat h^{\prime}}^{T}\vecfour{\quat h}$
for $\quat h,\,\quat h^{\prime}\in\mathbb{H}_{p}$.

\subsection{Constrained optimization algorithm\label{subsec:Constrained-optimization-algorithm}}

We control the translations of the instruments' tips using a centralized
kinematic control strategy \cite{marinhoUnifiedFrameworkTeleoperation2019}.
Let $\tilde{\quat t}_{i}\triangleq\tilde{\quat t}_{i}\left(\generalq\right)=\generalt-\quat t_{i\text{,d}}$
be the translation error between the current translation $\quat t_{\text{i}}\in\mathbb{H}_{p}$
and the desired translation $\quat t_{i\text{,d}}\in\mathbb{H}_{p}$
of the $i$-th robot's end effector, with $i\in\left\{ 1,2\right\} $.
Then, the desired control signal, $\myvec u=\begin{bmatrix}\myvec u_{1}^{T} & \myvec u_{2}^{T}\end{bmatrix}^{T}$,
is obtained as
\begin{align}
\myvec u\in\underset{\dotq}{\text{arg min}}\  & \beta\left(f_{\text{t,1}}+f_{\text{\ensuremath{\lambda},1}}\right)+\left(1-\beta\right)\left(f_{\text{t,2}}+f_{\text{\ensuremath{\lambda},2}}\right)\label{eq:constrained-optimization-algorithm}\\
\text{subject to} & \ \boldsymbol{W}\dotq\preceq\boldsymbol{w},\nonumber 
\end{align}
in which $\myvec q=\left[\begin{array}{cc}
\qsi^{T} & \qlg^{T}\end{array}\right]^{T}$, $f_{\text{t,i}}\triangleq\norm{\transjageneral\generalqdot+\eta\vecfour{\tilde{\quat t}_{i}}}_{2}^{2}$
are the cost functions related to the translation errors, $f_{\text{\ensuremath{\lambda},}i}\triangleq\lambda\norm{\generalqdot}_{2}^{2}$
are the cost functions related to the joint velocity norm, and $\transjageneral\in\mathbb{R}^{4\times n_{i}}$
are the translation Jacobians \cite{adornoDualPositionControl2010}
that satisfy $\vecfour{\dotgeneralt}=\transjageneral\generalqdot$.
In addition, $\eta\in\left(0,\infty\right)\subset\mathbb{R}$ is a
tunable gain, $\lambda\in[0,\infty)\subset\mathbb{R}$ is the damping
factor, and $\beta\in[0,\,1]\subset\mathbb{R}$ is a weight that defines
the priority between the two robots. The $r$ inequality constraints
$\boldsymbol{W}\dotq\preceq\boldsymbol{w}$, in which $\boldsymbol{W}\triangleq\boldsymbol{W}\left(\myvec q\right)\in\mathbb{R}^{r\times\left(n_{1}+n_{2}\right)}$
and $\boldsymbol{w}\triangleq\boldsymbol{w}\left(\myvec q\right)\in\mathbb{R}^{r}$,
are used to generate active constraints using VFIs \cite{marinhoDynamicActiveConstraints2019}.

\subsection{Vector-field-inequalities method\label{subsec:VFI-method}}

The VFI method \cite{marinhoDynamicActiveConstraints2019} uses signed
distance functions $d\triangleq d\left(\quat q,t\right)\in\mathbb{R}$
between two geometric primitives. The time-derivative of the distance
is
\begin{align*}
\dot{d}= & \underbrace{\partialfrac{d\left(\quat q,t\right)}{\quat q}}_{\boldsymbol{J}_{d}}\dot{\quat q}+\zeta\left(t\right)\text{,}
\end{align*}
where $\boldsymbol{J}_{d}\in\mathbb{R}^{1\times\left(\sijointnumber+\lgjointnumber\right)}$
is the distance Jacobian and $\zeta\left(t\right)=\dot{d}-\boldsymbol{J}_{d}\dot{\quat q}$
is the residual that contains the distance dynamics unrelated to $\dotq$.
Then, by using a safe distance $d_{\text{safe}}\triangleq d_{\text{safe}}\left(t\right)\in[0,\infty)$,
we define an error $\tilde{d}\triangleq\tilde{d}\left(\quat q,t\right)=d_{\text{safe}}-d$
to generate safe zones or $\tilde{d}\triangleq d-d_{\text{safe}}$
to generate restricted zones. With these definitions, and given $\eta_{d}\in[0,\infty)$,
the signed distance dynamics is constrained by $\dot{\tilde{d}}\geq-\eta_{d}\tilde{d}$
in both cases, which actively constrains the robot motion only in
the direction approaching the boundary between the primitives so that
the primitives do not collide. That is, the following constraint is
used to generate safe zones,
\begin{align}
\boldsymbol{J}_{d}\dot{\quat q}\leq & \eta_{d}\tilde{d}-\zeta_{\text{safe}}\left(t\right)\text{,}\label{eq:VFI-safe}
\end{align}
for $\zeta_{\text{safe}}\left(t\right)\triangleq\zeta\left(t\right)-\dot{d}_{\text{safe}}$.
Alternatively, restricted zones are generated by
\begin{align}
-\boldsymbol{J}_{d}\dot{\quat q}\leq & \eta_{d}\tilde{d}+\zeta_{\text{safe}}\left(t\right)\text{.}\label{eq:VFI-restrict}
\end{align}

\section{Robot-assisted vitreoretinal surgery\label{subsec:Autonomous-coordinated-control}}

In our previous work \cite{koyamaAutonomousCoordinatedControl2022},
we proposed a control strategy for robot-assisted vitreoretinal surgery
without orbital manipulation, including the autonomous coordinated
control of the light guide. This section summarizes the relevant parts
of \cite{koyamaAutonomousCoordinatedControl2022} used in this work.

Our control strategy relies on the constrained optimization problem
described in Section \ref{subsec:Constrained-optimization-algorithm}
as follows
\begin{align}
\myvec u\in\underset{\dotq}{\text{arg min}}\  & \beta\left(f_{\text{t,1}}+f_{\text{\ensuremath{\lambda},1}}\right)+(1-\beta)\left(f_{\text{t,2}}+f_{\text{\ensuremath{\lambda},2}}\right)\label{eq:constrained-optimization-algorithm-light}\\
\text{subject to} & \ \begin{bmatrix}\safeineqleft\\
\illuineqleft
\end{bmatrix}\dotq\preceq\begin{bmatrix}\safeineqright\\
\illuineqright
\end{bmatrix},\label{eq:previous_constraints}
\end{align}
where $\lgpd=0$, $\beta=0.99$, $\eta=140$, and $\lambda=0.001$.
The desired translation of the surgical needle, $\sipd$, is determined
by the operator or a predefined trajectory. The inequality constraints
are used to enforce the constraints to ensure safety, $\safeineqleft\dotq\leq\safeineqright$,
and the constraints for the autonomous control of the light guide,
$\illuineqleft\dotq\preceq\illuineqright$.

\begin{figure}[t]
\centering
\def\svgwidth{250pt} 
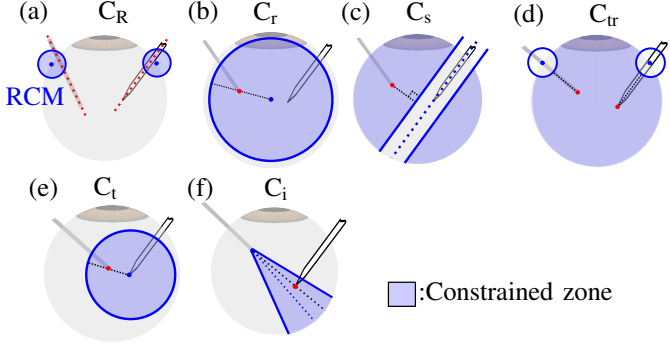

\caption{\label{fig:constraints} The illustrations of the geometric primitives
used to enforce the constraints for safety and the autonomous control
of the light guide.}
\end{figure}

The inequality constraints $\safeineqleft\dotq\preceq\safeineqright$
enforce the following constraints for safe vitreoretinal tasks
\begin{itemize}
\item $\crcm$: the shafts of the instruments must always pass through their
respective insertion points.
\item $\ceyeball$: the light guide's tip must never touch the retina.
\item $\cshaft$: the instruments' shafts must never collide with each other.
\item $\ctrocar$: the instruments' tips must always remain inside the eye.
\item $\cmicro$: the robots must never collide with the microscope.
\item $\crobot$: the robots must never collide with each other.
\item $\cjoint$: The robots' joint values must never exceed their limits.
\end{itemize}
Fig. \ref{fig:constraints}-(a), (b), (c), and (d) illustrate the
geometrical primitives we use to enforce $\crcm,\,\text{\ensuremath{\ceyeball}},\,\cshaft,\,\ctrocar$
using VFIs. Details are given in \cite[Section VIII]{koyamaAutonomousCoordinatedControl2022}.

The inequality constraints $\illuineqleft\dotq\preceq\illuineqright$
enforce the following constraints to autonomously control the light
guide
\begin{itemize}
\item $\ctip$: the needle's tip must be illuminated sufficiently.
\item $\cillu$: the needle's tip must be illuminated at all times.
\end{itemize}
Fig. \ref{fig:constraints}-(e), (f) illustrate the geometrical primitives
that enforce these constraints. Details are given in \cite[Section IX]{koyamaAutonomousCoordinatedControl2022}.

\section{Proposed orbital manipulation strategy}

In this section, we describe the main contribution of this work. Orbital
manipulation involves systematically moving the RCM positions of both
instruments in a coordinated way, such that the eye rotates about
its center. Orbital manipulation is frequently used by surgeons, for
instance, during the vitrectomy to check if there are no vitreous
cortex remnants. Given that the microscope and patient cannot be frequently
moved during the surgical procedure, orbital manipulation is the only
way to reach certain parts of the fundus and other relevant eye structures.

\subsection{Orbital manipulation VFI\label{subsec:Orbital-manipulation-VFI}}

\begin{figure}[t]
\centering
\def\svgwidth{200pt} 
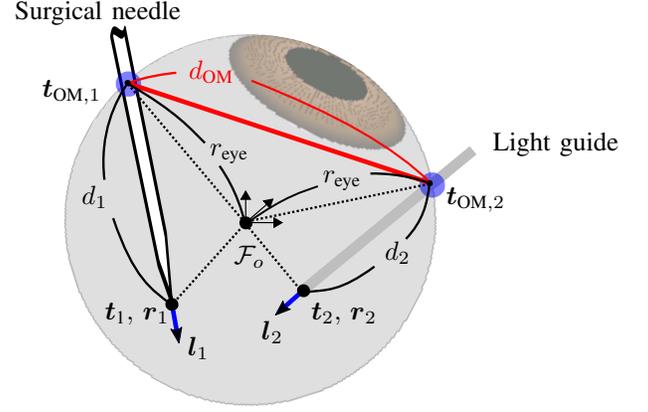

\caption{\label{fig:geometry-for-eyeball-manipulation} The geometric relationships
for the proposed orbital manipulation.}
\end{figure}

In geometrical terms, orbital manipulation is ensured by keeping the
relative position between the RCMs of each instrument while they otherwise
freely move. Let the distance between the RCMs be $d_{\orbital}\in\mathbb{R}^{+}$
as shown in Fig. \ref{fig:geometry-for-eyeball-manipulation}, we
constrain the squared distance\footnote{We use the squared distance since its time derivative, which we calculate
in Section \ref{subsec:new_rcm_jacobians}, is defined everywhere.} $D_{\orbital}\left(\myvec q\left(t\right)\right)\triangleq D_{\orbital}=d_{\orbital}^{2}$
as follows
\begin{gather}
-D_{\text{safe}}\leq D_{\orbital}-D_{\orbital,\text{init}}\leq D_{\text{safe}}\iff\nonumber \\
\underbrace{D_{\orbital}-\left(D_{\orbital,\text{init}}-D_{\text{safe}}\right)}_{\rcmdplus}\geq0,\ \underbrace{\left(D_{\orbital,\text{init}}+D_{\text{safe}}\right)-D_{\orbital}}_{\rcmdminus}\geq0\text{,}\label{eq:new_rcm_range}
\end{gather}
where $D_{\text{safe}}=0.5\,\mathrm{mm}$ and $D_{\orbital,\text{init}}=D_{\orbital}\left(\myvec q\left(t\right)\right)|_{t=0}$
are constants. Then, based on \eqref{eq:VFI-safe} and \eqref{eq:VFI-restrict},
these constraints in terms of joint velocities become
\begin{alignat}{1}
\left[\begin{array}{c}
-\jarcmdplus\\
\jarcmdminus
\end{array}\right]\dotq & \leq\eta_{\orbital}\left[\begin{array}{c}
\rcmdplus\\
\rcmdminus
\end{array}\right]\text{,}\label{eq:inequality_new_rcm}
\end{alignat}
where $\eta_{\orbital}=0.1$, and $\jarcmdplus\in\mathbb{R}^{1\times n}$
is the orbital manipulation Jacobian that relates the joint velocities
$\dotq$ to the time derivative of $\rcmdplus$ and $\rcmdminus$
in the form of $\dotrcmdplus=\jarcmdplus\dotq$ and $\dotrcmdminus=-\jarcmdplus\dotq$,
respectively. To enforce \eqref{eq:inequality_new_rcm}, we have to
find $D_{\orbital}$ and $\jarcmdplus$. We address this in Section
\ref{subsec:new_rcm_distance} and \ref{subsec:new_rcm_jacobians}.

\subsection{Orbital manipulation squared distance\label{subsec:new_rcm_distance}}

The goal of this section is to find the distance function $D_{\orbital}$
as a function of $\quat q$ to enforce \eqref{eq:inequality_new_rcm}.
We use the geometric relationships shown in Fig. \ref{fig:geometry-for-eyeball-manipulation}.
Let the world reference-frame $\worldreferenceframe$ be at the center
of the eyeball, and let the translations of the RCM points of both
instruments be $\generalrcmt\left(\generalq\right)\triangleq\generalrcmt\in\mathbb{H}_{p}\ \left(i=1,\,2\right)$.
Then, we have
\begin{align}
D_{\orbital}= & \norm{\sircmt-\lgrcmt}^{2}\text{.}\label{eq:new_rcm_squared_dis}
\end{align}

Next, we assume that, without loss of generality, $\generalline\left(\generalq\right)\triangleq\generalline\in\mathbb{S}^{3}\cap\mathbb{H}_{p}$
are the directions of the $z-$axes of the instruments pointing inside
the eye and given by
\begin{align}
\generalline= & \generalr\imk\conj{\left(\generalr\right)}\text{,}\label{eq:shaft_line}
\end{align}
where $\generalr\left(\generalq\right)\triangleq\generalr\in\mathbb{S}^{3}$
is the rotation of the instrument. Moreover, letting the distances
between the tips and the RCM positions be $\generallength\left(\generalq\right)\triangleq\generallength\in\mathbb{R}$,
we have
\begin{alignat}{1}
\generalrcmt & =\generalt-\generallength\generalline.\label{eq:rcm_translation}
\end{alignat}
Using the law of cosines, the radius of the eyeball $\eyeballradius\in\mathbb{R}^{+}-\left\{ 0\right\} $,
and \eqref{eq:rcm_translation}, we have
\begin{align*}
 & \begin{array}{cc}
\generallength^{2} & =\norm{\generalt}^{2}+\eyeballradius^{2}-2\dotmul{\generalt}{\generalrcmt}\\
 & =\eyeballradius^{2}-\norm{\generalt}^{2}+2\generallength\dotmul{\generalt}{\generalline}
\end{array}\\
\iff & \generallength^{2}-2\dotmul{\generalt}{\generalline}\generallength+\norm{\generalt}^{2}-\eyeballradius^{2}=0\\
\iff & \generallength=\dotmul{\generalt}{\generalline}\pm\sqrt{\dotmul{\generalt}{\generalline}^{2}-\norm{\generalt}^{2}+\eyeballradius^{2}}.
\end{align*}
Because of our definition of the direction for $\generalline$, only
the positive value is relevant.\footnote{The other value has an interesting mathematical meaning, because it
will be the other point in the sphere that satisfies this same equation
for $-\generalline$.} Therefore, we have
\begin{alignat}{1}
\generallength & =\dotmul{\generalt}{\generalline}+\underbrace{\sqrt{\dotmul{\generalt}{\generalline}^{2}-\norm{\generalt}^{2}+\eyeballradius^{2}}}_{h_{1}}\text{.}\label{eq:length}
\end{alignat}

\subsection{Orbital manipulation Jacobian\label{subsec:new_rcm_jacobians}}

Our next goal is to find the corresponding Jacobian $\jarcmdplus$
to enforce \eqref{eq:inequality_new_rcm}. Since the time derivatives
of $\rcmdplus$ and $\rcmdminus$ are $\dot{D}_{\orbital}$ and $-\dot{D}_{\orbital}$,
we can get $\jarcmdplus$ by finding the time derivative of $D_{\orbital}$
with respect to the joint velocities $\dotq$. From \eqref{eq:new_rcm_squared_dis},
we have
\begin{alignat}{1}
\dot{D}_{\orbital}= & 2\vecfour{\sircmt-\lgrcmt}^{T}\vecfour{\dotsircmt-\dotlgrcmt}\text{.}\label{eq:new_rcm_derivative1}
\end{alignat}

From \eqref{eq:rcm_translation}, the time derivative of $\generalrcmt\ \left(i=1,\,2\right)$
is
\begin{alignat}{1}
\vecfour{\dotgeneralrcmt} & =\transjageneral\generalqdot-\vecfour{\dotgenerallength\generalline+\generallength\dotgeneralline}.\label{eq:rcm_dot_1}
\end{alignat}

Then, we have to find $\dotgeneralline$ and $\dotgenerallength$.
From \eqref{eq:shaft_line}, we get
\begin{gather}
\vecfour{\dotgeneralline}=\underbrace{\left(\haminus{\imk\conj{\left(\generalr\right)}}+\haplus{\generalr\imk}\cfour\right)\rotationjageneral}_{\jageneralline}\generalqdot\text{,}\label{eq:dot_shaft_line}
\end{gather}
where $\rotationjageneral\in\mathbb{R}^{1\times n_{i}}$ is the rotation
Jacobian \cite{adornoDualPositionControl2010} that satisfies $\generaldotr=\rotationjageneral\generalqdot$.
As for $\dotgenerallength$, from \eqref{eq:length}, we have
\begin{alignat}{1}
\dotgenerallength & =\overbrace{\dotmul{\dotgeneralt}{\generalline}+\dotmul{\generalt}{\dotgeneralline}}^{h_{2}}+\dot{h}_{1}.\label{eq:d_dot}
\end{alignat}

Then, considering $h_{1}$ is a positive value,\footnote{Since the tip of the instrument is kept inside the eyeball, $\eyeballradius^{2}>\norm{\generalt}^{2}\iff\eyeballradius^{2}-\norm{\generalt}^{2}>0$.}
we get
\begin{alignat*}{1}
\dot{h_{1}} & =\frac{2\left(\dotmul{\dotgeneralt}{\generalline}+\dotmul{\generalt}{\dotgeneralline}\right)-2\dotmul{\generalt}{\dotgeneralt}}{h_{1}}\\
 & =\frac{1}{h_{1}}\left(2h_{2}-2\vecfour{\generalt}^{T}\vecfour{\dotgeneralt}\right),
\end{alignat*}
and, from \eqref{eq:dot_shaft_line},
\begin{alignat}{1}
h_{2}= & \underbrace{\left(\vecfour{\generalline}^{T}\transjageneral+\vecfour{\generalt}^{T}\jageneralline\right)}_{\mymatrix J_{h_{2}}}\generalqdot.\label{eq:h_2}
\end{alignat}
Therefore, we have
\begin{equation}
\dot{h_{1}}=\underbrace{\left(2\mymatrix J_{h_{2}}-2\vecfour{\generalt}^{T}\transjageneral\right)}_{\mymatrix J_{h_{1}}}\generalqdot.\label{eq:h_dot_1}
\end{equation}

We can now work backwards to find the corresponding Jacobian $\jarcmdplus$.
By substituting \eqref{eq:h_2} and \eqref{eq:h_dot_1} into \eqref{eq:d_dot},
we have
\begin{equation}
\dotgenerallength=\underbrace{\left(\mymatrix J_{h_{2}}+\mymatrix J_{h_{1}}\right)}_{\jagenerallength}\generalqdot\text{.}\label{eq:d_dot_2}
\end{equation}
Moreover, substituting \eqref{eq:dot_shaft_line} and \eqref{eq:d_dot_2}
into \eqref{eq:rcm_dot_1} results in
\begin{alignat}{1}
\vecfour{\dotgeneralrcmt} & =\underbrace{\left(\transjageneral-\vecfour{\generalline}\jagenerallength+\generallength\jageneralline\right)}_{\jarcmtgeneral}\generalqdot.\label{eq:rcm_dot_2}
\end{alignat}

Finally, from \eqref{eq:new_rcm_derivative1} and \eqref{eq:rcm_dot_2},
we find
\begin{alignat}{1}
\dot{D}_{\rcm}= & \underbrace{2\vecfour{\sircmt-\lgrcmt}^{T}\left[\begin{array}{cc}
\jasircmt & -\jalgrcmt\end{array}\right]}_{\jarcmdminus}\dotq\text{.}\label{eq:rcm_squared_d_dot}
\end{alignat}

\subsection{Constraints for safe orbital manipulation\label{subsec:Constraints-for-safe-orbital-manipulation}}

When the eye is autonomously rotated, safe limits must be enforced
to preserve the integrity of the eye muscles. To do so, we also propose
additional constraints.

\begin{figure}[t]
\centering
\def\svgwidth{250pt} 
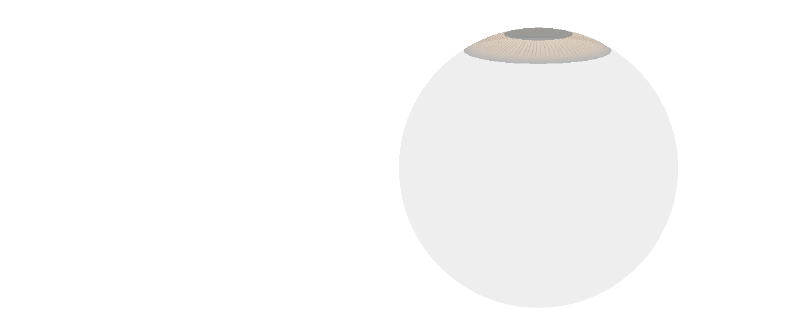

\caption{\label{fig:rotation_constraints} Geometrical primitives used to enforce
the constraints to limit the rotation of the eyeball, $\protect\crotationfirst$
and $\protect\crotationsecond$.}
\end{figure}

Fig. \ref{fig:rotation_constraints} shows the constraints $\crotationfirst$
and $\crotationsecond$. These constraints prevent the eyeball from
rotating beyond a certain angle. To this purpose, we constrain the
RCM positions $\sircmt$ and $\lgrcmt$ to be within a certain distance
from the planes $\rotationplanefirst$ and $\rotationplanesecond$.
The plane $\rotationplanefirst$ is the plane perpendicular to the
$x-$axis and contains the center of the eyeball. The plane $\rotationplanesecond$
is the plane perpendicular to the $z-$axis and $\rotplaned$ away
from the $xy-$plane. The distance $\rotplaned$ is half the radius
of the eyeball.

These constraints can be enforced using the signed distances and Jacobians
proposed in \cite[Eq. (57), (59)]{marinhoDynamicActiveConstraints2019}.
Let the signed distances and Jacobians between $\sircmt$ and $\rotationplanefirst,\,\rotationplanesecond$
be $\rotfirstdfirst,\,\rotseconddfirst\in\mathbb{R}$ and $\rotfirstjafirst,\,\rotsecondjafirst\in\mathbb{R}^{1\times\sijointnumber}$,
and the signed distances and Jacobians between $\lgrcmt$ and $\rotationplanefirst,\,\rotationplanesecond$
be $\rotfirstdsecond,\,\rotseconddsecond\in\mathbb{R}$ and $\rotfirstjasecond,\,\rotsecondjasecond\in\mathbb{R}^{1\times\lgjointnumber}$.
Then, to enforce $\crotation\ \left(j=1,\,2\right)$, we can use
\begin{alignat}{1}
\left[\begin{array}{cc}
-\rotgeneraljafirst & \boldsymbol{O}_{1\times\sijointnumber}\\
\boldsymbol{O}_{1\times\lgjointnumber} & -\rotgeneraljasecond
\end{array}\right]\dotq & \leq\eta_{\rotation}\left[\begin{array}{c}
\rotgeneraldfirst\\
\rotgeneraldsecond
\end{array}\right]\text{,}\label{eq:inequality_rotation}
\end{alignat}
where $\eta_{\rotation}=1.0$.

\section{Experiments}

We designed one simulation study and one experiment to evaluate our
proposed control strategy. First, we describe the simulation results
to evaluate the influence of the orbital manipulation on the manipulability
of our system. Then, we show experimental results to demonstrate the
feasibility of our method and the improvement in the microscopic field
of view enabled by the orbital manipulation when using the physical
robotic system.

\subsection{Setup}

The physical robotic setup shown in Fig. \ref{fig:surgical_system}
was used for the experiment. To observe the orbital manipulation,
the experiment was conducted without the face of the BionicEyE. For
the simulation study, this setup was replicated in CoppeliaSim (Coppelia
Robotics, Switzerland). Communication with the robot was enabled by
the SmartArmStack\footnote{https://github.com/SmartArmStack}. The
quaternion algebra and robot kinematics were implemented using DQ
Robotics \cite{adornoDQRoboticsLibrary2020} for Python3. The calibration
of the system and the registration of the initial RCM positions were
performed as described in \cite[Section X]{koyamaAutonomousCoordinatedControl2022}.

\subsection{Simulation: Evaluation of manipulability}

\begin{figure}[t]
\centering
\def\svgwidth{245pt} 
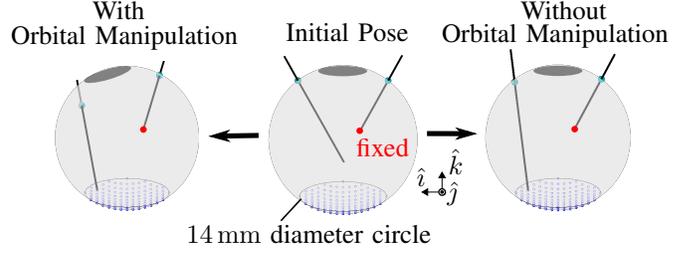

\caption{\label{fig:simulation} Positioning with and without the proposed
orbital manipulation in the simulation study.}
\end{figure}

\begin{figure}[t]
\centering
\def\svgwidth{245pt}
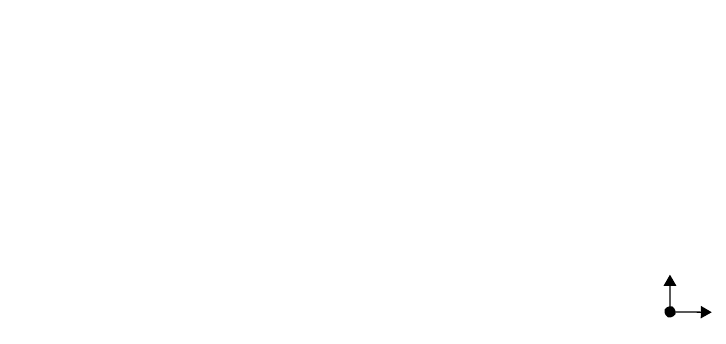

\caption{\label{fig:simulation_result} Manipulability measures \cite{yoshikawaManipulabilityRoboticMechanisms1985}
of $\protect\manijanewrcm$ and $\protect\manijaoldrcm$ calculated
at 149 points that cover a $14\,\mathrm{mm}$ diameter area around
the fundus in simulation.}
\end{figure}

\begin{figure*}
\centering
\def\svgwidth{500pt}
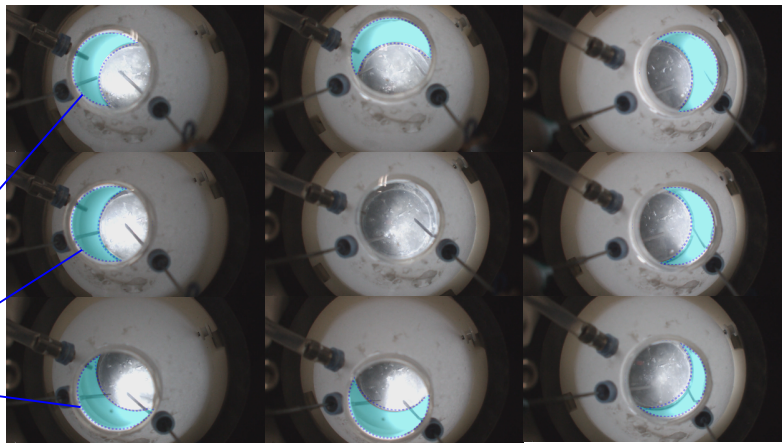

\caption{\label{fig:result_microscope} Comparison between conventional RCM-based
control and the proposed orbital manipulation strategy. (a) shows
the microscopic view with conventional RCM-based control. The eyeball
does not rotate, and the operator could see only the same regions
of the fundus of the eye model of the BionicEyE regardless of the
motion of the instruments. (b) shows the microscopic view with the
proposed autonomous orbital manipulation. Our strategy autonomously
rotates the eye model with respect to the motion of the instruments,
and the operator could observe the area around the fundus otherwise
unseen. The newly visible region was colored in light blue for easier
visualization.}
\end{figure*}

Inequality constraints, in general, can increase or reduce the manipulability
of the system in complex ways, and currently, there is no systematic
way of analyzing manipulability in this case. In this study, to assess
the impact of the proposed orbital manipulation on the manipulability
of the system, the manipulability measure proposed in \cite{yoshikawaManipulabilityRoboticMechanisms1985}
was used as a strict lower-bound to compare our proposed orbital manipulation
strategy and conventional fixed RCM-based control.

As shown in Fig. \ref{fig:simulation}, the procedure was as follows.
First, the surgical needle's tip was positioned to a point on the
fundus with and without enforcing the orbital manipulation. Then,
the manipulability measures after positioning were calculated and
compared. Positioning was conducted from the same initial pose to
149 positions covering a $14\,\mathrm{mm}$ diameter circle, which
is the double size of the usual target area. To make conditions equal,
the tip of the light guide was commanded to be still.

For a Jacobian $\manija$, the manipulability measure is calculated
as
\begin{align}
\manimeasure= & \sqrt{\text{det}\left|\manija\manija^{T}\right|},\label{eq:maipulability_measure}
\end{align}
which is related to the manipulability ellipsoid proposed in \cite{yoshikawaManipulabilityRoboticMechanisms1985}.
Compatible with the original definition, we used the following augmented
Jacobians
\begin{gather*}
\manijaoldrcm=\left[\begin{array}{cc}
\transjasi & \boldsymbol{O}_{3\times\lgjointnumber}\\
\boldsymbol{O}_{3\times\sijointnumber} & \transjalg\\
\jasircmt & \boldsymbol{O}_{1\times\lgjointnumber}\\
\boldsymbol{O}_{1\times\sijointnumber} & \jalgrcmt
\end{array}\right]\text{ and }\manijanewrcm=\left[\begin{array}{c}
\begin{array}{cc}
\transjasi & \boldsymbol{O}_{3\times\lgjointnumber}\end{array}\\
\begin{array}{cc}
\boldsymbol{O}_{3\times\sijointnumber} & \transjalg\end{array}\\
\simulationrcmjanew
\end{array}\right],
\end{gather*}
for the fixed-RCM evaluation and the orbital manipulation evaluation,
respectively. Note that $\transjageneral$ are the translation Jacobians,
and $\jarcmtgeneral$ are as found in \eqref{eq:rcm_dot_2}. Moreover,
$\simulationrcmjanew$ is the proposed orbital manipulation Jacobian
that satisfies $\simulationrcmjanew\dotq=\totalderivative{\norm{\sircmt-\lgrcmt}}t$.
That is because it is well known that, to calculate the manipulability,
we need to unify the units of the augmented Jacobian. Hence, by letting
$\quat h_{3}=\norm{\sircmt-\lgrcmt}>0$, the Jacobian $\simulationrcmjanew$
can be calculated as follows.
\begin{align*}
\dot{\quat h}_{3}= & \underbrace{\frac{1}{\quat h_{3}}\vecfour{\sircmt-\lgrcmt}^{T}\left[\begin{array}{cc}
\jasircmt & -\jalgrcmt\end{array}\right]}_{\simulationrcmjanew}\dotq.
\end{align*}

\subsubsection{Results and discussion}

Fig. \ref{fig:simulation_result}-(a) shows the manipulability measures
of $\manijanewrcm$ and $\manijaoldrcm$ calculated at the 149 points.
The figure shows that the orbital manipulation provided higher values
of the manipulability measure. The manipulability measure of $\manijaoldrcm$
was always zero because one of the singular values related to the
fixed RCM constraint was always zero.

Fig. \ref{fig:simulation_result}-(b) shows the manipulability measures
of $\manijanewrcm$ at each point. The manipulability measures of
the points away from the RCMs were higher than those of the points
near the RCMs. This means that the tip of the surgical needle can
smoothly access the points far from the RCMs by moving the RCMs.

\subsection{Experiment: Evaluation of real-world feasibility\label{subsec:Experiment}}

This experiment was conducted to show the feasibility of the proposed
method and compare it with a conventional approach. In this experiment,
the tip of the surgical needle was tele-operatively controlled using
an input device (Touch, 3D Systems, USA). The operator arbitrarily
moved the tip parallel to the image plane of the microscope in two
modes, with the conventional fixed-RCM and the proposed orbital manipulation
strategy. The field-of-view of the microscope was larger than in actual
surgery for a proper workspace analysis.

\subsubsection{Results and discussion}

As shown in Fig. \ref{fig:result_microscope}-(a), with the conventional
approach, the operator could see only the same region of the fundus
of the eye model. On the other hand, as shown in Fig. \ref{fig:result_microscope}-(b),
the orbital manipulation enabled the operator to observe a larger
area around the fundus. We also confirmed that the proposed constraints
were satisfied, and the orbital manipulation was enforced autonomously
with respect to the motion of the tip of the surgical needle.

Moreover, the light guide autonomously followed the tip of the surgical
needle during the experiments, and the constraints for safety and
the automation of the light guide were also satisfied.

Lastly, it cannot be understated that the proposed orbital manipulation
based on VFIs comes at very little cost. In fact, the quadratic programming
solver has one less constraint to solve with respect to having two
fixed RCMs. In addition, there was no change in task description nor
increase in the number of control parameters. Enacting limits on the
orbital manipulation increases the number of design parameters, but
those are intrinsic to the eye and require no specific tuning.

\section{Conclusion}

In this paper, we proposed a new control strategy for orbital manipulation.
To achieve this, we derived a new distance function and its corresponding
Jacobian to keep the relative position between the RCMs using VFIs.
In a simulation and an experiment, we showed that orbital manipulation
increased the manipulability of the system and enabled the operator
to observe a larger area around the fundus.

Future works include the investigation of the sclera force during
orbital manipulation, the consideration of the eyeball pose, and the
evaluation of the feasibility of vitreoretinal tasks otherwise impossible
without orbital manipulation.

\bibliographystyle{IEEEtran}
\bibliography{ICRA2023}

\ifdefined\DeclarePrefChars\DeclarePrefChars{'’-}\else\fi
\begin{thebibliography}{10}
\providecommand{\url}[1]{#1}
\csname url@samestyle\endcsname
\providecommand{\newblock}{\relax}
\providecommand{\bibinfo}[2]{#2}
\providecommand{\BIBentrySTDinterwordspacing}{\spaceskip=0pt\relax}
\providecommand{\BIBentryALTinterwordstretchfactor}{4}
\providecommand{\BIBentryALTinterwordspacing}{\spaceskip=\fontdimen2\font plus
\BIBentryALTinterwordstretchfactor\fontdimen3\font minus
  \fontdimen4\font\relax}
\providecommand{\BIBforeignlanguage}[2]{{%
\expandafter\ifx\csname l@#1\endcsname\relax
\typeout{** WARNING: IEEEtran.bst: No hyphenation pattern has been}%
\typeout{** loaded for the language `#1'. Using the pattern for}%
\typeout{** the default language instead.}%
\else
\language=\csname l@#1\endcsname
\fi
#2}}
\providecommand{\BIBdecl}{\relax}
\BIBdecl

\bibitem{singhPhysiologicalTremorAmplitude2002}
\BIBentryALTinterwordspacing
S.~Singh and C.~Riviere, ``Physiological tremor amplitude during retinal
  microsurgery,'' in \emph{Proceedings of the {{IEEE}} 28th {{Annual Northeast
  Bioengineering Conference}}}.\hskip 1em plus 0.5em minus 0.4em\relax {IEEE},
  pp. 171--172. [Online]. Available:
  \url{https://ieeexplore.ieee.org/document/999520/}
\BIBentrySTDinterwordspacing

\bibitem{iordachitaRoboticAssistanceIntraocular2022}
\BIBentryALTinterwordspacing
I.~i.~Iordachita, M.~D. de~Smet, G.~Naus, M.~Mitsuishi, and C.~N. Riviere,
  ``Robotic {{Assistance}} for {{Intraocular Microsurgery}}: {{Challenges}} and
  {{Perspectives}},'' \emph{Proceedings of the IEEE}, pp. 1--16. [Online].
  Available: \url{https://ieeexplore.ieee.org/document/9771085/}
\BIBentrySTDinterwordspacing

\bibitem{maclachlanMicronActivelyStabilized2012}
\BIBentryALTinterwordspacing
R.~A. MacLachlan, B.~C. Becker, J.~C. Tabares, G.~W. Podnar, L.~A. Lobes, and
  C.~N. Riviere, ``Micron: {{An Actively Stabilized Handheld Tool}} for
  {{Microsurgery}},'' \emph{IEEE Transactions on Robotics}, vol.~28, no.~1, pp.
  195--212. [Online]. Available:
  \url{https://ieeexplore.ieee.org/document/6084852/}
\BIBentrySTDinterwordspacing

\bibitem{kimDesignControlFully2021}
E.~Kim, I.~Choi, and S.~Yang, ``Design and {{Control}} of {{Fully Handheld
  Microsurgical Robot}} for {{Active Tremor Cancellation}},'' in \emph{2021
  {{International Conference}} on {{Robotics}} and {{Automation}} ({{ICRA}})},
  p.~7.

\bibitem{uneriNewSteadyhandEye2010}
\BIBentryALTinterwordspacing
A.~Uneri, M.~A. Balicki, J.~Handa, P.~Gehlbach, R.~H. Taylor, and
  I.~Iordachita, ``New steady-hand {{Eye Robot}} with micro-force sensing for
  vitreoretinal surgery,'' in \emph{2010 3rd {{IEEE RAS}} \& {{EMBS
  International Conference}} on {{Biomedical Robotics}} and
  {{Biomechatronics}}}.\hskip 1em plus 0.5em minus 0.4em\relax {IEEE}, pp.
  814--819. [Online]. Available:
  \url{http://ieeexplore.ieee.org/document/5625991/}
\BIBentrySTDinterwordspacing

\bibitem{gijbelsClinicallyApplicableRobotic2016}
A.~Gijbels, K.~Willekens, L.~Esteveny, P.~Stalmans, D.~Reynaerts, and
  E.~Vander~Poorten, ``Towards a clinically applicable robotic assistance
  system for retinal vein cannulation,'' in \emph{2016 6th {{IEEE International
  Conference}} on {{Biomedical Robotics}} and {{Biomechatronics}}
  ({{BioRob}})}, pp. 284--291.

\bibitem{weiDesignTheoreticalEvaluation2007}
W.~Wei, R.~Goldman, N.~Simaan, H.~Fine, and S.~Chang, ``Design and
  {{Theoretical Evaluation}} of {{Micro-Surgical Manipulators}} for {{Orbital
  Manipulation}} and {{Intraocular Dexterity}},'' in \emph{Proceedings 2007
  {{IEEE International Conference}} on {{Robotics}} and {{Automation}}}, pp.
  3389--3395.

\bibitem{nasseriIntroductionNewRobot2013}
M.~A. Nasseri, M.~Eder, S.~Nair, E.~C. Dean, M.~Maier, D.~Zapp, C.~P. Lohmann,
  and A.~Knoll, ``The introduction of a new robot for assistance in ophthalmic
  surgery,'' in \emph{2013 35th {{Annual International Conference}} of the
  {{IEEE Engineering}} in {{Medicine}} and {{Biology Society}} ({{EMBC}})}, pp.
  5682--5685.

\bibitem{wilsonIntraocularRoboticInterventional2018}
\BIBentryALTinterwordspacing
J.~T. Wilson, M.~J. Gerber, S.~W. Prince, C.-W. Chen, S.~D. Schwartz, J.-P.
  Hubschman, and T.-C. Tsao, ``Intraocular robotic interventional surgical
  system ({{IRISS}}): {{Mechanical}} design, evaluation, and master-slave
  manipulation: {{Intraocular}} robotic interventional surgical system
  ({{IRISS}}),'' \emph{The International Journal of Medical Robotics and
  Computer Assisted Surgery}, vol.~14, no.~1, p. e1842. [Online]. Available:
  \url{http://doi.wiley.com/10.1002/rcs.1842}
\BIBentrySTDinterwordspacing

\bibitem{edwardsFirstinhumanStudySafety2018a}
\BIBentryALTinterwordspacing
T.~L. Edwards, K.~Xue, H.~C.~M. Meenink, M.~J. Beelen, G.~J.~L. Naus, M.~P.
  Simunovic, M.~Latasiewicz, A.~D. Farmery, M.~D. de~Smet, and R.~E. MacLaren,
  ``First-in-human study of the safety and viability of intraocular robotic
  surgery,'' \emph{Nature Biomedical Engineering}, vol.~2, no.~9, pp. 649--656.
  [Online]. Available: \url{https://www.nature.com/articles/s41551-018-0248-4}
\BIBentrySTDinterwordspacing

\bibitem{gijbelsInHumanRobotAssistedRetinal2018}
\BIBentryALTinterwordspacing
A.~Gijbels, J.~Smits, L.~Schoevaerdts, K.~Willekens, E.~B. Vander~Poorten,
  P.~Stalmans, and D.~Reynaerts, ``In-{{Human Robot-Assisted Retinal Vein
  Cannulation}}, {{A World First}},'' \emph{Annals of Biomedical Engineering},
  vol.~46, no.~10, pp. 1676--1685. [Online]. Available:
  \url{http://link.springer.com/10.1007/s10439-018-2053-3}
\BIBentrySTDinterwordspacing

\bibitem{marinhoSmartArmIntegrationValidation2020}
\BIBentryALTinterwordspacing
M.~M. Marinho, K.~Harada, A.~Morita, and M.~Mitsuishi, ``{{SmartArm}}:
  {{Integration}} and validation of a versatile surgical robotic system for
  constrained workspaces,'' \emph{The International Journal of Medical Robotics
  and Computer Assisted Surgery}, vol.~16, no.~2, apr 2020. [Online].
  Available: \url{https://onlinelibrary.wiley.com/doi/abs/10.1002/rcs.2053}
\BIBentrySTDinterwordspacing

\bibitem{weiweiPerformanceEvaluationMultiarm2009}
\BIBentryALTinterwordspacing
{Wei Wei}, R.~Goldman, H.~Fine, {Stanley Chang}, and N.~Simaan, ``Performance
  {{Evaluation}} for {{Multi-arm Manipulation}} of {{Hollow Suspended
  Organs}},'' \emph{IEEE Transactions on Robotics}, vol.~25, no.~1, pp.
  147--157. [Online]. Available:
  \url{http://ieeexplore.ieee.org/document/4694099/}
\BIBentrySTDinterwordspacing

\bibitem{yuDesignCalibrationPreliminary2013}
\BIBentryALTinterwordspacing
H.~Yu, J.-H. Shen, K.~M. Joos, and N.~Simaan, ``Design, calibration and
  preliminary testing of a robotic telemanipulator for {{OCT}} guided retinal
  surgery,'' in \emph{2013 {{IEEE International Conference}} on {{Robotics}}
  and {{Automation}}}.\hskip 1em plus 0.5em minus 0.4em\relax {IEEE}, pp.
  225--231. [Online]. Available:
  \url{http://ieeexplore.ieee.org/document/6630580/}
\BIBentrySTDinterwordspacing

\bibitem{marinhoUnifiedFrameworkTeleoperation2019}
\BIBentryALTinterwordspacing
M.~M. Marinho, B.~V. Adorno, K.~Harada, K.~Deie, A.~Deguet, P.~Kazanzides,
  R.~H. Taylor, and M.~Mitsuishi, ``A {{Unified Framework}} for the
  {{Teleoperation}} of {{Surgical Robots}} in {{Constrained Workspaces}},'' in
  \emph{2019 {{International Conference}} on {{Robotics}} and {{Automation}}
  ({{ICRA}})}.\hskip 1em plus 0.5em minus 0.4em\relax {IEEE}, pp. 2721--2727.
  [Online]. Available: \url{https://ieeexplore.ieee.org/document/8794363/}
\BIBentrySTDinterwordspacing

\bibitem{tomikiUseGeneralpurposeSeriallink2017}
Y.~Tomiki, M.~M. Marinho, Y.~Kurose, K.~Harada, and M.~Mitsuishi, ``On the use
  of general-purpose serial-link manipulators in eye surgery,'' in \emph{2017
  14th {{International Conference}} on {{Ubiquitous Robots}} and {{Ambient
  Intelligence}} ({{URAI}})}, pp. 540--541.

\bibitem{koyamaAutonomousCoordinatedControl2022}
Y.~Koyama, M.~M. Marinho, M.~Mitsuishi, and K.~Harada, ``Autonomous
  {{Coordinated Control}} of the {{Light Guide}} for {{Positioning}} in
  {{Vitreoretinal Surgery}},'' \emph{{IEEE} Transactions on Medical Robotics
  and Bionics}, vol.~4, no.~1, pp. 156--171, feb 2022.

\bibitem{heAutomaticLightPipe2020}
\BIBentryALTinterwordspacing
C.~He, E.~Yang, N.~Patel, A.~Ebrahimi, M.~Shahbazi, P.~Gehlbach, and
  I.~Iordachita, ``Automatic {{Light Pipe Actuating System}} for {{Bimanual
  Robot-Assisted Retinal Surgery}},'' \emph{IEEE/ASME Transactions on
  Mechatronics}, vol.~25, no.~6, pp. 2846--2857. [Online]. Available:
  \url{https://ieeexplore.ieee.org/document/9099104/}
\BIBentrySTDinterwordspacing

\bibitem{kimAutonomousEyeSurgery2021}
\BIBentryALTinterwordspacing
J.~W. Kim, P.~Zhang, P.~Gehlbach, I.~Iordachita, and M.~Kobilarov, ``Towards
  {{Autonomous Eye Surgery}} by {{Combining Deep Imitation Learning}} with
  {{Optimal Control}},'' \emph{Proceedings of machine learning research}, vol.
  155, pp. 2347--2358. [Online]. Available:
  \url{https://www.ncbi.nlm.nih.gov/pmc/articles/PMC8549631/}
\BIBentrySTDinterwordspacing

\bibitem{shinSemiAutomatedExtractionLens2021}
C.~Shin, M.~J. Gerber, Y.-H. Lee, M.~Rodriguez, S.~A. Pedram, J.-P. Hubschman,
  T.-C. Tsao, and J.~Rosen, ``Semi-{{Automated Extraction}} of {{Lens Fragments
  Via}} a {{Surgical Robot Using Semantic Segmentation}} of {{OCT Images With
  Deep Learning}} - {{Experimental Results}} in {{Ex Vivo Animal Model}},''
  \emph{IEEE Robotics and Automation Letters}, vol.~6, no.~3, pp. 5261--5268.

\bibitem{dehghaniColibriDocEyeInHandAutonomous2022}
S.~Dehghani, M.~Sommersperger, J.~Yang, M.~Salehi, B.~Busam, K.~Huang,
  P.~Gehlbach, I.~I. Iordachita, N.~Navab, and M.~A. Nasseri, ``{{ColibriDoc}}:
  {{An Eye-In-Hand Autonomous Trocar Docking System}},'' in \emph{2021
  {{International Conference}} on {{Robotics}} and {{Automation}} ({{ICRA}})},
  p.~7.

\bibitem{leeDualRedundantArm1989}
S.~Lee, ``Dual redundant arm configuration optimization with task-oriented dual
  arm manipulability,'' \emph{IEEE Transactions on Robotics and Automation},
  vol.~5, no.~1, pp. 78--97.

\bibitem{wenKinematicManipulabilityGeneral1999a}
J.-Y. Wen and L.~Wilfinger, ``Kinematic manipulability of general constrained
  rigid multibody systems,'' \emph{IEEE Transactions on Robotics and
  Automation}, vol.~15, no.~3, pp. 558--567.

\bibitem{alambeigiSemiautonomousCryoablationKidney2018}
\BIBentryALTinterwordspacing
F.~Alambeigi, Z.~Wang, Y.-h. Liu, R.~H. Taylor, and M.~Armand, ``Toward
  {{Semi-autonomous Cryoablation}} of {{Kidney Tumors}} via {{Model-Independent
  Deformable Tissue Manipulation Technique}},'' \emph{Annals of Biomedical
  Engineering}, vol.~46, no.~10, pp. 1650--1662. [Online]. Available:
  \url{http://link.springer.com/10.1007/s10439-018-2074-y}
\BIBentrySTDinterwordspacing

\bibitem{marinhoDynamicActiveConstraints2019}
\BIBentryALTinterwordspacing
M.~M. Marinho, B.~V. Adorno, K.~Harada, and M.~Mitsuishi, ``Dynamic {{Active
  Constraints}} for {{Surgical Robots Using Vector-Field Inequalities}},''
  \emph{{IEEE} Transactions on Robotics}, vol.~35, no.~5, pp. 1166--1185, oct
  2019. [Online]. Available:
  \url{https://ieeexplore.ieee.org/document/8742769/}
\BIBentrySTDinterwordspacing

\bibitem{omataSurgicalSimulatorPeeling2018a}
\BIBentryALTinterwordspacing
S.~Omata, Y.~Someya, S.~Adachi, T.~Masuda, T.~Hayakawa, K.~Harada,
  M.~Mitsuishi, K.~Totsuka, F.~Araki, M.~Takao, M.~Aihara, and F.~Arai, ``A
  surgical simulator for peeling the inner limiting membrane during wet
  conditions,'' \emph{PLOS ONE}, vol.~13, no.~5, p. e0196131. [Online].
  Available: \url{https://dx.plos.org/10.1371/journal.pone.0196131}
\BIBentrySTDinterwordspacing

\bibitem{adornoRobotKinematicModeling2017}
B.~V. Adorno, \emph{Robot {{Kinematic Modeling}} and {{Control Based}} on
  {{Dual Quaternion Algebra}} - {{Part I}}: {{Fundamentals}}.}

\bibitem{adornoDualPositionControl2010}
B.~V. Adorno, P.~Fraisse, and S.~Druon, ``Dual position control strategies
  using the cooperative dual task-space framework,'' in \emph{2010
  {{IEEE}}/{{RSJ International Conference}} on {{Intelligent Robots}} and
  {{Systems}}}, pp. 3955--3960.

\bibitem{adornoDQRoboticsLibrary2020}
\BIBentryALTinterwordspacing
B.~V. Adorno and M.~Marques~Marinho, ``{{DQ Robotics}}: {{A Library}} for
  {{Robot Modeling}} and {{Control}},'' \emph{{IEEE} Robotics and Automation
  Magazine}, vol.~28, no.~3, pp. 102--116, sep 2021. [Online]. Available:
  \url{https://ieeexplore.ieee.org/document/9136790/}
\BIBentrySTDinterwordspacing

\bibitem{yoshikawaManipulabilityRoboticMechanisms1985}
\BIBentryALTinterwordspacing
T.~Yoshikawa, ``Manipulability of {{Robotic Mechanisms}},'' \emph{The
  International Journal of Robotics Research}, vol.~4, no.~2, pp. 3--9.
  [Online]. Available: \url{https://doi.org/10.1177/027836498500400201}
\BIBentrySTDinterwordspacing

\end{thebibliography}

\end{document}